\documentclass{article}

\usepackage{PRIMEarxiv}
\usepackage{makecell}
\usepackage[utf8]{inputenc} 
\usepackage[T1]{fontenc}    
\usepackage{hyperref}       
\usepackage{url}            
\usepackage{booktabs}       
\usepackage{amsfonts}       
\usepackage{nicefrac}       
\usepackage{microtype}      
\usepackage{lipsum}
\usepackage{fancyhdr}       
\usepackage{graphicx}       
\graphicspath{{media/}}     

\usepackage{multirow}
\usepackage{subcaption}  
\usepackage{amsmath}

\title{TIBR4D: Tracing-Guided Iterative Boundary Refinement for Efficient 4D Gaussian Segmentation
}

\author{
  He Wu, Xia Yan, Yanghui Xu, Liegang Xia, Jiazhou Chen*\\
  Zhejiang University of Technology, China\\
  \texttt{cjz@zjut.edu.cn} \\
}

\begin{document}
\maketitle


\begin{abstract}
Object-level segmentation in dynamic 4D Gaussian scenes remains challenging due to complex motion, occlusions, and ambiguous boundaries. 
In this paper, we present an efficient learning-free 4D Gaussian segmentation framework that lifts video segmentation masks to 4D spaces, whose core is a two-stage iterative boundary refinement, TIBR4D. 
The first stage is an Iterative Gaussian Instance Tracing (IGIT) at the temporal segment level. It progressively refines Gaussian-to-instance probabilities through iterative tracing, and extracts corresponding Gaussian point clouds that better handle occlusions and preserve completeness of object structures compared to existing one-shot threshold-based methods. 
The second stage is a frame-wise Gaussian Rendering Range Control (RCC) via suppressing highly uncertain Gaussians near object boundaries while retaining their core contributions for more accurate boundaries.
Furthermore, a temporal segmentation merging strategy is proposed for IGIT to balance identity consistency and dynamic awareness. Longer segments enforce stronger multi-frame constraints for stable identities, while shorter segments allow identity changes to be captured promptly.
Experiments on HyperNeRF and Neu3D demonstrate that our method produces accurate object Gaussian point clouds with clearer boundaries and higher efficiency compared to SOTA methods.
\end{abstract}


\section{Introduction}
\label{sec:intro}

This paper focuses on efficient and precise object-level segmentation in dynamic scenes. Real-world environments are inherently dynamic, with objects undergoing motion, deformation, and frequent interactions over time, making it particularly challenging to achieve efficient, reliable, and dynamic-aware object segmentation. On the other hand, such a capability is crucial for accurate perception, long-horizon planning, and robust interaction, and is relevant to a wide range of applications, including autonomous driving, robotics, and augmented or virtual reality.

The rapid advancement of novel view synthesis techniques \cite{mildenhall2021nerf,kerbl20233d,barron2021mip,yu2024mip} has significantly improved scene modeling capabilities. Neural Radiance Fields (NeRF) \cite{mildenhall2021nerf} demonstrate remarkable expressive power for static scenes, and have been extended to incorporate semantic information \cite{cen2023segment,kim2024garfield,kundu2022panoptic,kerr2023lerf,kobayashi2022decomposing} or support dynamic 4D scene representations \cite{li2022neural,park2021nerfies,pumarola2021d,fang2022fast,park2021hypernerf}. However, NeRF-based approaches rely on implicit scene representations, which incur substantial computational overhead during training and rendering, limiting their applicability to downstream tasks. In contrast, 3D Gaussian Splatting (3DGS) \cite{kerbl20233d} employs a learnable explicit representation, achieving faster rendering and improved interpretability. Building on this foundation, some approaches incorporate language features into 3DGS to support open-vocabulary querying and semantic interaction \cite{qin2024langsplat,shi2024language,wu2024opengaussian,jun2025dr,li20254d}, while others focus on instance-level segmentation, aiming to decompose complex scenes into semantically meaningful objects \cite{cen2025segment,ye2024gaussian,dou2024learning,lyu2024gaga,shen2025trace3d}.  Meanwhile, Gaussian-based representations have been extended to the dynamic scene \cite{wu20244d,yang2024deformable,li2024spacetime,wu2025orientation}. These advances lay a strong foundation for exploring 4D Gaussian scene segmentation; however, it is still challenging to achieve efficient, reliable, and dynamics-aware object segmentation in 4D Gaussian scenarios.
\begin{figure}[t]
\centering
\begin{minipage}[t]{0.32\textwidth}
    \centering
    \includegraphics[width=\linewidth]{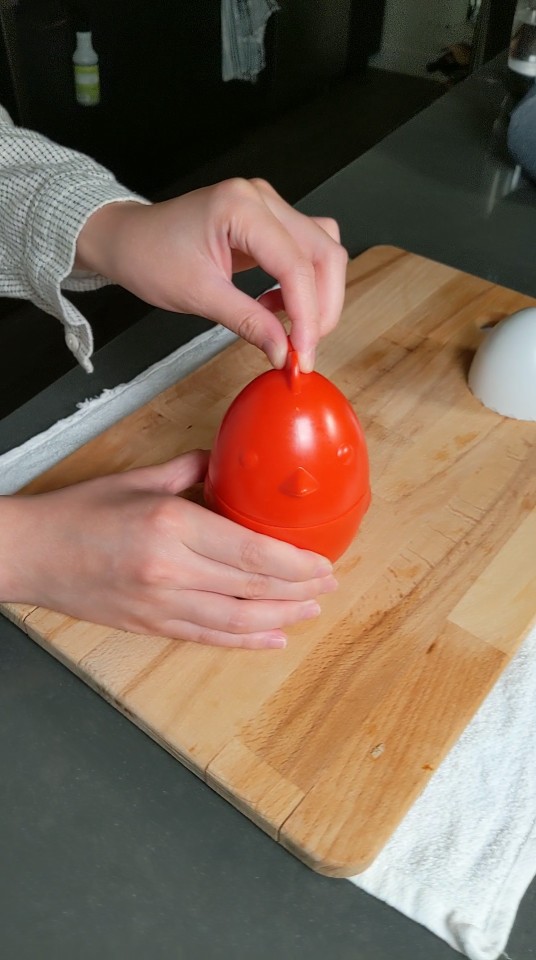}
    \small Reference Image
\end{minipage}
\begin{minipage}[t]{0.64\textwidth}
\centering
\begin{minipage}[t]{0.24\linewidth}
    \centering
    \includegraphics[width=\linewidth]{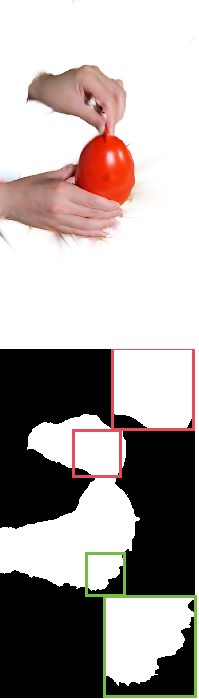}\\
    \small SA4D
\end{minipage}
\begin{minipage}[t]{0.24\linewidth}
    \centering
    \includegraphics[width=\linewidth]{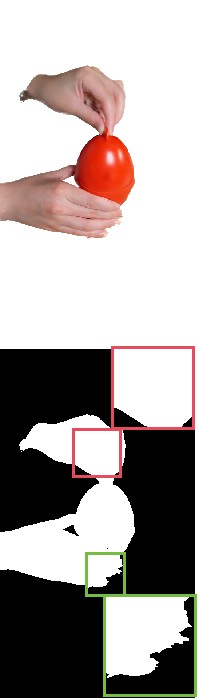}\\
    \small SADG
\end{minipage}
\begin{minipage}[t]{0.24\linewidth}
    \centering
    \includegraphics[width=\linewidth]{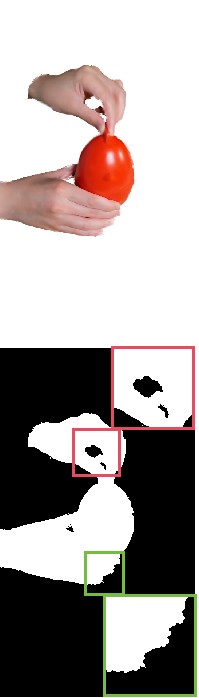}\\
    \small Ours
\end{minipage}
\begin{minipage}[t]{0.24\linewidth}
    \centering
    \includegraphics[width=\linewidth]{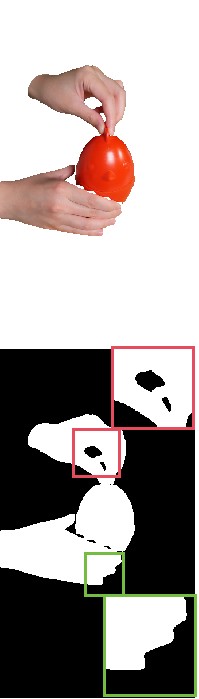}\\
    \small GT
\end{minipage}
\end{minipage}
\caption{
\textbf{Comparison with prior works SA4D \cite{ji2024segment} and SADG \cite{li2024sadg}}. Previous methods often suffer from overflow at object boundaries. Benefiting from two convergent iterative stages, our method enables fast extraction of target objects while effectively suppressing floating Gaussians and boundary leakage, resulting in more accurate segmentation boundaries.
}
\label{fig:fig1}
\end{figure}

Representative challenges in 4D Gaussian segmentation cover three key aspects, including: 1) achieving \textit{accurate instance boundaries} under occlusions and complex motion, 2) balancing \textit{dynamic scene expressiveness} with segmentation stability, and 3) maintaining practical \textit{efficiency}. Existing 4D Gaussian segmentation methods generally follow two distinct paradigms: (i) tracking-based approaches that lift 2D video segmentation labels into the 4D space, producing time-varying Gaussian identity features, and (ii) tracking-free approaches that focus on learning consistent and stable Gaussian identity features via contrastive learning. 

Tracking-based methods, represented by SA4D \cite{ji2024segment} and CIF \cite{wu2025consistent}, predict the identity features per-frame for each Gaussian using MLPs. These methods typically require shorter feature learning time and exhibit stronger responsiveness to dynamic appearance and structural changes. Nevertheless, frequent temporal updates of Gaussian identities often lead to instability during object extraction, manifested as noticeable Gaussian flickering artifacts. Additionally, due to the complex occlusion relationships in dynamic scenes, they commonly suffer from floating points and boundary leakage, which significantly degrade the quality of object-level 4D segmentation.

Tracking-free methods, such as SADG \cite{li2024sadg}, learn fixed identity features for deformable Gaussians via contrastive learning, which alleviates visual artifices caused by label inconsistency in tracking–based supervision. Nevertheless, maintaining fixed Gaussian identities over time limits its ability to handle dynamic scenes where Gaussian identities may evolve. Split4D \cite{hu2025split4d} further introduces linear motion modeling for Gaussian primitives and adopts a streaming sampling strategy to enable Gaussian reuse, partially mitigating the limitations of fixed identities. However, identity ambiguities may still arise in scenes with complex occlusions. Moreover, boundary leakage of segmented objects is still not fully addressed in these methods, and contrastive learning–based methods generally incur substantial feature learning overhead, which affects overall efficiency in practice. 

In this paper, we propose a learning-free segmentation framework that efficiently lifts 2D video segmentation labels into the 4D space by iteratively tracing the probabilities of Gaussians belonging to each object.
To address challenges arising from floating points and boundary leakage, our framework consists of two convergent iterative refinement stages: Iterative Gaussian Instance Tracing (IGIT) on the temporal segment level and frame-wise Gaussian Rendering Range Control (RRC). 
IGIT removes most floating points by iteratively refining instance assignments, enabling better handling of occluded Gaussians and preserving more complete object structures compared to existing one-shot threshold-based extraction. 
RRC performs fine-grained, frame-wise intra-Gaussian control over Gaussian rendering ranges, suppressing the rendering ranges of highly uncertain Gaussians near object boundaries, while preserving their core contributions around the Gaussian centers, resulting in more accurate segmentation boundaries. 
It is worth noting that IGIT is conducted on temporal segments instead of the whole video or frame-wise, to balance identity consistency and dynamic awareness. Longer temporal segments impose stronger multi-frame constraints, leading to more stable and consistent Gaussian point clouds, while independent processing across segments allows identity changes to be captured promptly. Thus, we additionally introduce a temporal segmentation merging strategy that progressively merges segments during the iterative tracing process to automatically determine an appropriate temporal granularity.

In summary, our contributions are:
\begin{itemize}
    \item We present a learning-free object-level 4D Gaussian segmentation framework with tracing-guided iterative boundary refinement. Gaussians are iteratively traced and segmented at the temporal segment level.
    \item We propose a frame-wise Gaussian rendering range control strategy for fine-grained intra-Gaussian segmentation. This strategy is convergent and allows iterative refinement for precise object boundaries.
    \item Experimental results on the HyperNeRF \cite{park2021hypernerf} and Neu3D \cite{li2022neural} datasets show that our method achieves more accurate rendered masks for dynamic Gaussian segmentation compared to existing approaches.
\end{itemize}

\section{Related Work}
\label{sec:related}

\textbf{4D Scene Representation}: Existing 4D scene reconstruction methods often build upon established 3D reconstruction techniques and extend them to handle dynamic scenes. The introduction of Neural Radiance Fields (NeRFs) \cite{mildenhall2021nerf} and their extensions \cite{barron2021mip,barron2023zip} has brought significant advancements to 3D scene reconstruction, enabling full-scene modeling through fully connected networks over implicit 3D spaces and allowing photorealistic image synthesis from novel viewpoints. Numerous works have further extended NeRF to dynamic scene reconstruction \cite{li2022neural,park2021nerfies,pumarola2021d,fang2022fast,park2021hypernerf}. However, NeRF-based approaches are typically limited by the implicit nature of their representations, which incur high computational costs and restrict their applicability in downstream tasks. In contrast, 3D Gaussian Splatting (3DGS) \cite{kerbl20233d} and subsequent methods that optimize reconstruction performance \cite{yu2024mip,lu2024scaffold,zeng2025frequency} adopt an explicit, learnable representation of scenes, achieving more efficient real-time rendering. Building on this capability, many approaches have incorporated 3DGS representations into dynamic scene reconstruction \cite{wu20244d,yang2024deformable,li2024spacetime,wu2025orientation}, leveraging techniques such as deformations and trajectory modeling to facilitate downstream tasks, including dynamic scene understanding and interaction.

\textbf{3D Segmentation}: Recent advances in vision foundation models, such as SAM \cite{kirillov2023segment}, DINO \cite{caron2021emerging}, DINOv2 \cite{oquab2023dinov2}, CLIP \cite{radford2021learning}, and foundation model-based frameworks (e.g., DEVA \cite{cheng2023tracking}), have provided good opportunities for semantic understanding in 3D and 4D scenes. A number of NeRF-based approaches \cite{cen2023segment,kim2024garfield,kundu2022panoptic,kerr2023lerf,kobayashi2022decomposing} have explored lifting semantic information extracted from vision foundation models to the 3D scene level, enabling semantic-aware neural radiance field representations through supervision from 2D semantic cues. With the emergence of 3D Gaussian Splatting (3DGS) \cite{kerbl20233d}, its explicit and learnable scene representation offers fast rendering and improved interpretability, making it particularly suitable for semantic scene understanding. Consequently, many recent works have adopted Gaussian-based representations to incorporate semantic information into scene modeling. Some methods \cite{qin2024langsplat,shi2024language,wu2024opengaussian,jun2025dr,li20254d} focus on semantic segmentation by combining Gaussian scene representations with language features to enable open-vocabulary querying, while others target instance-level segmentation \cite{cen2025segment,ye2024gaussian,dou2024learning,lyu2024gaga,shen2025trace3d}. These instance-level approaches typically leverage 2D masks or video trackers to provide direct mask-based supervision, or employ contrastive loss to learn a 3D-consistent feature field across views. Notably, Trace3D \cite{shen2025trace3d} introduces Gaussian Instance Tracing (GIT) to efficiently compute a weight matrix for each Gaussian and further proposes a GIT-guided adaptive density control strategy to split and prune ambiguous Gaussians during training. This design enables fast lifting of 2D segmentation results to 3D Gaussian scenes and leads to more accurate segmentation boundaries. However, most existing methods are designed for static 3D scenes and lack efficient mechanisms for fast and robust instance segmentation in dynamic 4D Gaussian scenes.

\textbf{Dynamic Gaussian-based Instance Segmentation}: In prior work on instance segmentation for 4D Gaussian scenes, DGD \cite{labe2024dgddynamic3dgaussians} extends the 3DGS \cite{kerbl20233d} rasterization pipeline to render per-Gaussian semantic features and supervises them using features extracted from DINOv2 \cite{oquab2023dinov2} or CLIP \cite{radford2021learning}. However, the high dimensionality of these features (384/512) introduces non-negligible training overhead. SADG \cite{li2024sadg} adopts a contrastive learning strategy by assigning each Gaussian primitive a set of learnable identity features that remain fixed across time in dynamic scenes, but it cannot handle cases where Gaussian identity information changes over time. Split4D \cite{hu2025split4d} addresses large-motion scenarios by endowing Gaussians with linear motion over time to improve the temporal consistency of Gaussian identities, but handling complex occlusions and boundary overflow remains challenging. 

SA4D \cite{ji2024segment} and CIF \cite{wu2025consistent} leverage video tracking techniques \cite{cheng2023tracking} to obtain temporally consistent instance labels, lift 2D segmentation results into the 4D domain using a 4D feature field, and assume that Gaussian identity information varies over time. SA4D \cite{ji2024segment} employs an MLP to predict dynamically changing Gaussian features at each time step, but it does not explicitly account for occlusions, often resulting in noticeable artifacts in the segmented objects. Additionally, the time-varying Gaussian identity features may cause Gaussian flickering during object extraction. CIF \cite{wu2025consistent}, in addition to learning Gaussian features, further requires estimating spatial occupancy probabilities and refines object boundaries through a Gaussian splitting strategy, which introduces a higher learning burden compared to SA4D \cite{ji2024segment}. 

However, these methods rely on learning Gaussian features for 4D segmentation, which incurs substantial training costs and hinders fast segmentation of target objects from 2D instance cues. Moreover, these features, learned from the original scene, are directly used to extract target point clouds. This necessitates empirically setting high score thresholds to identify objects of interest and often leads to difficulty in balancing the suppression of boundary leakage with the preservation of object content. %
To address these challenges, we introduce an efficient learning-free object-level 4D Gaussian segmentation framework. It employs two tracing-based iterative refinement stages to achieve precise object boundaries while maintaining the integrity of the object content.
\section{Method}
\label{sec:method}

\begin{figure}
  \centering
  \includegraphics[width=\textwidth]{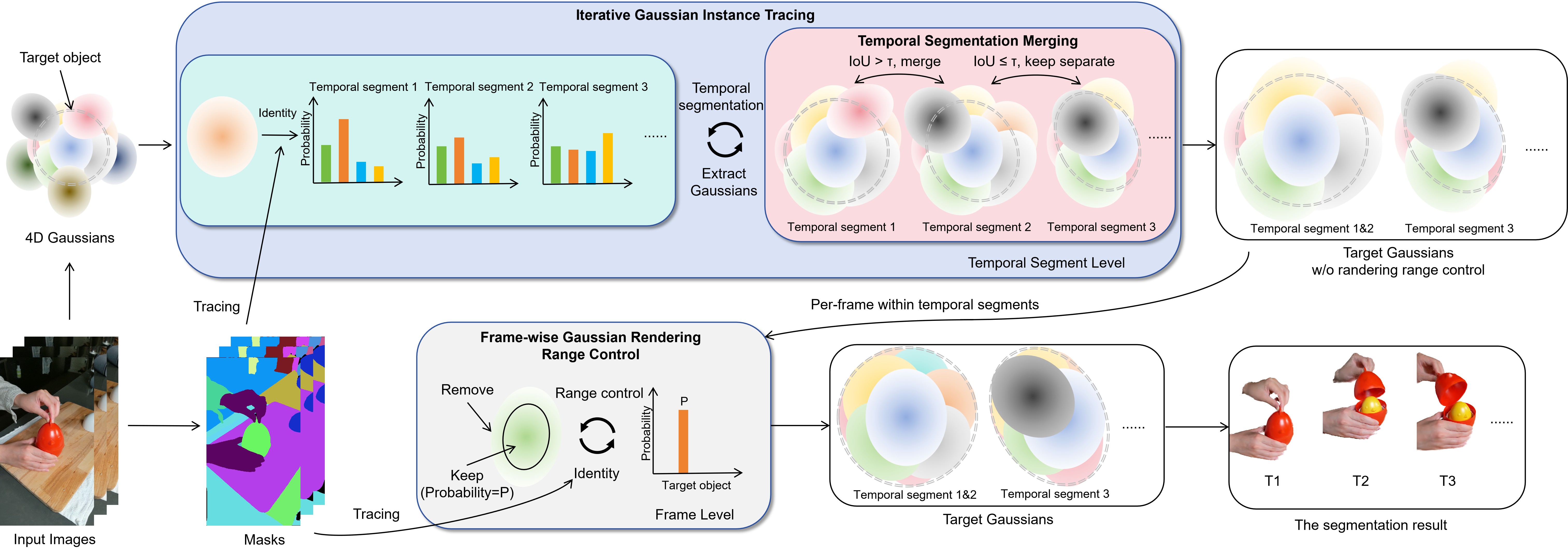}
  \caption{
  \textbf{Overview}. Our learning-free object segmentation framework of 4D Gaussian scenes consists of two iterative stages: iterative Gaussian instance tracing (Sec. \ref{sec:IGIT}) with temporal segmentation (Sec. \ref{sec:TSM}), and frame-wise Gaussian rendering range control (Sec. \ref{sec:RRC}). In the first stage, we iteratively perform Gaussian Instance Tracing within temporal segments to remove erroneous Gaussians and merge adjacent segments when their extracted Gaussian point clouds become consistent. In the second stage, we progressively truncate the outer regions of each projected 2D Gaussian and retain only its most reliable central contribution. After these two convergent stages, we obtain a clean Gaussian point cloud for the target object.
  }
  \label{fig:overview}
\end{figure}
Figure~\ref{fig:overview} shows the overall framework of the proposed 4D Gaussian segmentation method, which consists of two iterative stages. In the first stage, Gaussian Instance Tracing \cite{shen2025trace3d} is iteratively applied together with Gaussian extraction to obtain a relatively accurate and clean Gaussian point cloud of the target objects (Sec. \ref{sec:IGIT}). This process is performed over temporal segments, enabling identity changes of Gaussians in dynamic scenes to be captured while maintaining segmentation consistency. A temporal segmentation merging strategy is further introduced to determine the appropriate temporal segmentation granularity (Sec. \ref{sec:TSM}). In the second stage, the probability of each Gaussian belonging to the target object is leveraged as guidance for frame-wise Gaussian rendering range control. Based on these probabilities, instance tracing is iteratively performed on each frame to progressively refine a reliable rendering range for each Gaussian distribution(Sec. \ref{sec:RRC}).

\subsection{Iterative Gaussian Instance Tracing}
\label{sec:IGIT}
We adopt Gaussian Instance Tracing (GIT) \cite{shen2025trace3d} to lift 2D instance segmentation results to the 3D Gaussian point cloud and estimate the probability that each Gaussian belongs to different object instances. Specifically, we first obtain the 2D instance segmentation mask $M_t$ of the frame $t$ using DEVA \cite{cheng2023tracking}. Then, for each training view $v_{t}$, the Gaussian scene is projected onto the corresponding image plane. Based on the instance IDs of the pixels covered by each Gaussian and the contribution of the Gaussian distribution to these pixels, we compute a weight matrix that represents the association between Gaussians and instance IDs in the current view. This weight matrix is denoted as $W_{t}\in \mathbb{R}^{N\times K}$, where $N$ is the number of Gaussians in the scene and $K$ is the number of unique instance IDs produced by DEVA \cite{cheng2023tracking} for the current scene. The weight of a Gaussian $G_{i}$ belonging to instance ID $k$ under view $v_{t}$ is defined as:
\begin{equation}
W_t(G_i,k)
=
\sum_{(u,v)}
\mathbb{I}\big[ M_t(u,v)=k \big]\,
o_i(t)\,
G_i^{2D}(u,v,t)\,
T_i(u,v,t)
\end{equation}
\noindent where $o_{i}(t)$ denotes the opacity of the $i^{th}$ Gaussian at the frame $t$, $G_{i}^{2D}(u,v,t)$ represents the value of the 2D Gaussian distribution evaluated at pixel location $(u,v,t)$, $T_i(u,v,t) = \prod_{j=1}^{i-1} \bigl(1 - o_j(t)\,G_j^{2D}(u,v,t)\bigr)$ which represents the accumulated transmittance, $\mathbb{I}\big[\cdot\big]$ denotes the indicator function, $\mathbb{I}\big[ M_t(u,v) = k \big]$ equals to 1 when $M_t(u,v) = k$ is true (i.e., the pixel $(u,v,t)$ belongs to object $k$), and otherwise it equals to 0. After summing the identity weight matrices of all Gaussians across all views and applying normalization, we obtain the probability matrix $P\in[0,1]^{N\times K}$, which represents the probability of each Gaussian belonging to each instance identity over all time steps in the current scene:
\begin{equation}
P_{i,:}=\frac{\textstyle\sum_{t}{W_t(G_i,:)}}{\textstyle||{\textstyle\sum_{t}{W_t(G_i,:)}}||}
\end{equation}
For the target object with instance ID $k_{0}$, we define a binary Gaussian mask $M_{k_0}^G \in \{0,1\}^N$, $M_{k_0}^G = \mathbb{I}[\, k_0 = \arg\max_j P_{:,j} \,]$. Using this mask, a coarse Gaussian point cloud corresponding to the target object can be extracted.

However, a single pass of Gaussian Instance Tracing is insufficient to extract an accurate Gaussian point cloud of the target objects, since many Gaussians are partially or fully occluded in the original 4D Gaussian scene. Although the visible portions of these Gaussians are primarily associated with the target objects, once they are extracted, the previously occluded parts may become visible, leading to changes in the instance membership probabilities of these Gaussians. In practice, this phenomenon manifests as numerous incorrectly segmented floating Gaussians in the resulting point cloud. Previous 4D Gaussian instance segmentation methods \cite{ji2024segment,li2024sadg,hu2025split4d}, typically compute identity features based on the original 4D Gaussian scene. However, these features are not well-suited for the already segmented Gaussian point cloud, where occlusion relationships among Gaussians have changed. As a result, these methods typically perform only a single step of point cloud extraction and rely on manually set high score thresholds to remove floating Gaussians. However, such threshold selection is heuristic and lacks robustness, often requiring scene-specific tuning. Moreover, this single-step inference scheme struggles to eliminate all floating Gaussians while preserving the object content.

A straightforward idea is to ignore the occlusion from foreground Gaussians when computing Gaussian identity weights. However, this strategy would cause background Gaussians to be incorrectly preserved, which in turn degrades the segmentation quality. Therefore, accounting for Gaussian visibility remains necessary in our formulation.

Benefiting from the aforementioned learning-free formulation, we can efficiently perform multiple rounds of Gaussian identity estimation. For the Gaussian point cloud obtained after an initial segmentation, we re-project the 2D instance segmentation results onto the extracted Gaussians, thereby re-estimating the probabilities under the updated occlusion relationships. This enables a more accurate refinement of Gaussian identities. By iteratively repeating this process, we progressively obtain a cleaner and more reliable Gaussian point cloud:
\begin{equation}
W_t(G_i,k)
=
\sum_{(u,v)}
\mathbb{I}\big[ M_t(u,v) = k \big] \,
o_i(t) \,
G_i^{2D}(u,v,t) \,
T_i(u,v,t, M_{k_0}^G) \,
\times M_{k_0}^G
\end{equation}
\noindent where $T_{i}(u,v,t,M_{k_0}^G) = \prod_{j<i} \big( 1 - o_{j}(t)\, G_{j}^{2D}(u,v,t)\, \times M_{k_0}^G \big)$. It means that Gaussians that have already been determined not to belong to the target object do not participate in further identity or occlusion computations.

In our experiments, we set the number of iterations of Instance Tracing over all training views by default to 20. One iteration consists of applying instance tracing once to each training view in the scene. This ensures that the extracted Gaussian point cloud is sufficiently clean. The number of iterations can be set quite high because the method is convergent; in later iterations, the extracted Gaussian point cloud exhibits minimal change. In practice, for most scenes, only 5 iterations are sufficient to achieve very good segmentation results. A detailed discussion of the number of iterations and convergence behavior is provided in Sec. \ref{sec:Comparison}.

\subsection{Temporal Segmentation}
\label{sec:TSM}
In a 4D Gaussian scene, the identity of each Gaussian may have a large temporal variation. Taking the \textit{cut-lemon1} scene as an example, Gaussians at the cut part of the lemon shrink, while Gaussians at the new location deform to form the cut lemon slice. This phenomenon is not physically correct, but is visually acceptable. However, in such a case with varying Gaussian identities, 4D Gaussian segmentation methods with fixed Gaussian identities tend to produce visual artifacts and content loss, such as SADG~\cite{li2024sadg}. Instead, methods such as \cite{ji2024segment,wu2025consistent} predict Gaussian identities at each time step using MLPs, and thus can effectively capture temporal variations in Gaussian identities. However, for relatively stable Gaussian scenes, per-frame prediction may introduce undesirable flickering of certain Gaussians. 
To avoid the constancy of global-fixed identities and the instability of per-frame identity prediction, our IGIT is applied at the level of temporal segments. Longer segments preserve stable Gaussian identity features, while independence across segments allows identity changes to be captured promptly.
Therefore, it is necessary to adopt an appropriate temporal segmentation strategy to determine at which time points in the dynamic scene should be partitioned into temporal segments. 

Specifically, we first divide the scene into temporal segments based on each training view frame. For each segment $s \in \{1, \dots, S\}$, we perform the probability computation described in Sec. \ref{sec:IGIT} to extract the target object Gaussian point cloud. Then, for the Gaussian point clouds $M_{k_0}^{G,s}$ extracted from segment s, we compute the Intersection-over-Union (IoU) with the point cloud from the adjacent segment $M_{k_0}^{G,s+1}$. If the IoU exceeds a threshold $\tau$, the two temporal segments are merged:

\begin{equation}
\mathrm{IoU}\big(M_{k_0}^{G,s},\, M_{k_0}^{G,s+1}\big) > \tau 
\end{equation}

We set $\tau=0.5$, which works well in all of our experiments. In practice, users can adjust $\tau$ according to the temporal variation of Gaussian identities. Subsequently, the Gaussian point cloud segmentation is updated based on the merged temporal segments, producing updated masks $M_{k_0}^{G,s}$, and the merging process is repeated. This iterative temporal segment merging is executed in synchrony with the iterative Gaussian projection and segmentation operation described in Sec.~\ref{sec:IGIT}. In scenes where Gaussian information is relatively stable, the temporal merging typically results in a single segment. In contrast, for scenes with more complex changes in Gaussian identities, such as \textit{cut-lemon1}, multiple temporal segments are necessary.

\subsection{Frame-wise Gaussian Rendering Range Control}
\label{sec:RRC}
After the procedures described in Sec. \ref{sec:IGIT} and Sec. \ref{sec:TSM}, we obtain relatively clean Gaussian point clouds for each temporal segment. However, two types of Gaussians still adversely affect the final mask rendering quality. The first type consists of Gaussians located near object boundaries or Gaussians that occupy an excessively large screen space. Although these Gaussians are assigned high probabilities of belonging to the target object within a temporal segment, a substantial portion of their support lies outside the true object boundaries, which negatively impacts the rendered masks. The second type includes Gaussians whose identity assignments are unstable within a temporal segment. While the temporal segment merging strategy in Sec. \ref{sec:TSM} encourages overall identity consistency of the extracted Gaussian point clouds, it cannot effectively handle individual Gaussians whose identities change within a segment. Naively removing these two types of Gaussians on a per-frame basis is undesirable, as it leads to overly sparse Gaussian point clouds and exacerbates large-scale Gaussian flickering, ultimately degrading the rendering quality.

Therefore, we propose to control the rendering range of each Gaussian on a frame-wise basis, retaining only the most central and informative portion of each Gaussian’s contribution. This strategy aims to produce cleaner segmentation boundaries while avoiding overly sparse Gaussian point clouds and minimizing adverse effects on the rendering results. Similar to the procedure in Sec. \ref{sec:IGIT}, we project the 2D segmentation masks onto the Gaussians for each training frame, yielding the probability $p_{i,t,k_0}$, that Gaussian $i$ belongs to the target object $k_{0}$ at time $t$. We then use this probability $p_{i,t,k_0}$ as an initial reference threshold $r_{i,t,k_0}$ for controlling the Gaussian rendering range, which is incorporated into both instance tracing and rendering. During instance tracing and rendering, for each 2D Gaussian projected onto the screen, we render only the central region whose cumulative probability mass equals $r_{i,t,k_0}$. The outer region is excluded from both rendering and occlusion computations.

In addition, unlike the IGIT stage, occlusion effects from foreground Gaussians on the current Gaussian (denoted as $T_{i}(u,v,t,r,M_{k_0}^{G,s})$) are not considered during instance tracing at this stage. One reason is that the influence of Gaussian visibility on segmentation has already been incorporated in the IGIT stage. Gaussians that were incorrectly retained due to ignoring visibility have been effectively removed. More importantly, the control of the rendering range is performed independently for the projected 2D distribution of each Gaussian. By excluding occlusion relationships, the estimated probability $p_{i,t,k_0}$ of a Gaussian belonging to the target object becomes more stable, which in turn facilitates better convergence of the rendering range threshold $r_{i,t,k_0}$ and helps avoid overly restrictive truncation in certain cases:
\begin{equation}
\label{eq:RRC_W}
W_t(G_i, k)
=
\sum_{(u,v)}
\mathbb{I}\big[ M_t(u,v) = k \big] \,
o_i(t) \,
G_i^{2D}(u,v,t) \,
M_{k_0}^{G,s} \,
\times \mathbb{I}\big[ G_i^{2D}(u,v,t) > (1 - r_{i,t,k_0}) \big] 
\end{equation}
\begin{equation}
\label{eq:RRC_P}
p_{i,t,k_0} = \frac{W_t(G_i, k_0)}{\textstyle||{\textstyle{W_t(G_i,:)}}||}
\end{equation}
\noindent where $M_{k_0}^{G,s}$ denotes the Gaussian point cloud segmented by the procedures described in Secs. \ref{sec:IGIT} and \ref{sec:TSM} for time segment $s$, to which the current frame $t$ belongs. The indicator function $\mathbb{I}\big[ G_i^{2D}(u,v,t) > (1 - r_{i,t,k_0}) \big]$ selects the central region of each projected 2D Gaussian $G_i^{2D}$, ensuring that only this region contributes to rendering. Accordingly, the rendering formulation for the segmentation result of object $k_{0}$ is modified as follows:
\begin{equation}
c(u, v, t)
=
\sum_{i=1} 
o_i(t) \,
G_i^{2D}(u,v,t) \,
T_i(u,v,t,r, M_{k_0}^{G,s}) \,
M_{k_0}^{G,s} \,
\times \mathbb{I}\big[ G_i^{2D}(u,v,t) > (1 - r_{i,t,k_0}) \big]
\end{equation}
\noindent where $T_i(u,v,t,r,M_{k_0}^{G,s}) 
= \prod_{j<i} 
\Big( 1 - o_j(t)\, G_j^{2D}(u,v,t)\, 
M_{k_0}^{G,s} \,
\times \mathbb{I}\big[ G_j^{2D}(u,v,t) > (1 - r_{j,t,k_0}) \big] \Big)$, which represents the accumulated transmittance within only the central region of each projected 2D Gaussian $G_i^{2D}$ constrained by the threshold $r_{i,t,k_0}$.

Ideally, the removed portion of each Gaussian, corresponding to cumulative probability $1 - r_{i,t,k_0}$, would not belong to the target object. In practice, however, since we retain only the central region corresponding to the cumulative probability $r_{i,t,k_0}$ for each Gaussian, this approach is an approximation and does not guarantee that the removed portion is entirely outside the object $k_{0}$. Consequently, a single application of rendering range control may still result in substantial boundary leakage. Benefiting from the convergent nature of this range control operation, multiple iterations can be performed to obtain more accurate Gaussian truncation thresholds.

In the initialization stage, we use the probability $p_{i,t,k_0}$ obtained from untruncated Gaussians before any range control to initialize the threshold $r_{i,t,k_0}$ for subsequent iterative refinement. Then we iteratively perform Instance Tracing over all training frames to compute the new probability $p_{i,t,k_0}$ that each truncated Gaussian belongs to the target object, following Eqs. \ref{eq:RRC_W} and \ref{eq:RRC_P}, and use these probabilities to update $r_{i,t,k_0}$. Since this iterative process recomputes $p_{i,t,k_0}$ based on the Gaussian regions that truncated according to $r_{i,t,k_0}$, the update of $r_{i,t,k_0}$ is defined in a multiplicative form:
\begin{equation}
r_{i,t,k_0}^{Iteration+1} = r_{i,t,k_0}^{Iteration}  \times p_{i,t,k_0}^{Iteration}
\end{equation}

In our experiments, we set the default number of iterations to 20. Due to the convergent nature of the method, when the number of iterations is sufficiently large, the probability $p_{i,t,k_0}$ for each Gaussian at each frame tends to either 1 (indicating that the Gaussian is fully within the rendering range of the target object) or 0 (indicating that the Gaussian is completely outside, for example, when its distribution center lies just beyond the object boundary in a given frame). At this point, the value of $r_{i,t,k_0}$ has effectively converged.
In the iteration stage, we obtain segmentation results with comparatively well-defined object boundaries. We store the segmentation results of the target object $k_{0}$ in two matrices: $M_{k_0}^G \in \{0,1\}^{N \times S}$, $M_{k_0}^r \in [0,1]^{N \times T}$.

\section{Experiments}
\label{sec:experi}

In this section, we first introduce the experimental setup for the dynamic Gaussian scene segmentation task, including the datasets, benchmarks, evaluation metrics, and baselines (Sec. \ref{sec:setup}). We then compare our method with recent competitive methods for 4D Gaussian instance segmentation on the HyperNeRF \cite{park2021hypernerf} and Neu3D \cite{li2022neural} datasets (Sec. \ref{sec:Comparison}). We conduct ablation studies to evaluate the impact of different components of our method on the experimental results (Sec. \ref{sec:Ablation}). Finally, we discuss the limitations of our method  (Sec. \ref{sec:limit}).

\subsection{Experimental Setup}
\label{sec:setup}
\textbf{Datasets}: We evaluate our method on two widely-used datasets: HyperNeRF \cite{park2021hypernerf} and Neu3D \cite{li2022neural}. The HyperNeR dataset is captured using a single moving monocular camera, observing dynamic scenes with diverse human-object interactions~\cite{park2021hypernerf}. The Neu3D dataset consists of multi-view videos recorded with a rig of 18–21 cameras at 30 FPS, depicting a person performing cooking tasks in a kitchen~\cite{li2022neural}. Following SADG \cite{li2024sadg}, we conduct experiments on these HyperNeRF \cite{park2021hypernerf} scenes: \textit{americano}, \textit{split-cookie}, \textit{oven-mitts}, \textit{espresso}, \textit{chickchicken}, \textit{hand1-dense-v2}, \textit{torchocolate}, \textit{slice-banana}, \textit{keyboard}, and \textit{cut-lemon1}; and on these Neu3D~\cite{li2022neural} scenes: \textit{coffee\_martini}, \textit{cook\_spinach}, \textit{cut\_roasted\_beef}, \textit{flame\_steak}, and \textit{sear\_steak}. We adopt the benchmark annotations provided in SADG~\cite{li2024sadg} for evaluation.

\textbf{Evaluation Metrics}: To evaluate the quality of the segmented masks, we follow the evaluation protocol of SA4D \cite{ji2024segment} and SADG \cite{li2024sadg}, i.e., adopting Mean Intersection over Union (mIoU) and Mean Pixel Accuracy (mAcc) as quantitative metrics for segmentation performance. In addition, we measure the time required by three different methods to learn instance identity features and extract target objects across scenes in the HyperNeRF \cite{park2021hypernerf} and Neu3D \cite{li2022neural} datasets. This timing evaluation excludes the reconstruction of dynamic Gaussian scenes and only accounts for the time spent on Gaussian identity feature learning and target object point cloud extraction. All experiments are conducted on a single NVIDIA RTX 3090 GPU.

\textbf{Baselines}: We compare our method with recent competitive methods for 4D Gaussian instance segmentation SA4D \cite{ji2024segment} and SADG \cite{li2024sadg}. For SA4D~\cite{ji2024segment} and our method, we conduct experiments using the same reconstructed 4D Gaussian point clouds and the same 2D segmentation results generated from DEVA \cite{cheng2023tracking}, ensuring a fair comparison. For SADG~\cite{li2024sadg}, we follow the original training settings provided in the official implementation and obtain results that are close to those reported in the paper. For multi-view datasets Neu3D \cite{li2022neural}, we select the training view that is closest to the test views to generate 2D segmentation masks for both SA4D~\cite{ji2024segment} and our method, i.e., the \textit{cam05} scene.

\begin{table}
 \caption{
 Quantitative comparison of our method’s performance against SA4D \cite{ji2024segment} and SADG \cite{li2024sadg} on the HyperNeRF \cite{park2021hypernerf} dataset. Our method outperforms these SOTA methods on most of scenes (both mIoU and mAcc in \%).
 }
  \centering
  \begin{tabular}{lllllllllllll}
    \toprule
    \multirow{2}{*}{Methods}  & \multicolumn{2}{c}{americano} & \multicolumn{2}{c}{chickchicken} & \multicolumn{2}{c}{cut-lemon1} & \multicolumn{2}{c}{espresso} & \multicolumn{2}{c}{hand} & \multicolumn{2}{c}{keyboard}                   \\
        &  mIoU      &  mAcc     &  mIoU      &  mAcc    &  mIoU      &  mAcc    &  mIoU      &  mAcc    &  mIoU      &  mAcc    &  mIoU      &  mAcc  \\
    \midrule
    SA4D & 84.08 & 99.43 &87.93  &96.91  &83.40  &96.55  &73.08  &98.64  &\textbf{91.77}  &98.63  &84.97  &97.33 \\
    SADG &78.64 &99.09  &93.89  &98.56  &\textbf{87.60}  &\textbf{97.64}  &69.66  &98.43  &91.46  &\textbf{98.72}  &87.35  &97.85\\
    Our &\textbf{86.08}  &\textbf{99.55}  &\textbf{95.22}  &\textbf{98.89}  &85.25  &97.20  &\textbf{90.40}  &\textbf{99.62}  &90.78  &98.48  &\textbf{94.12}  &\textbf{99.07}\\
    \midrule
    \multirow{2}{*}{Methods}      & \multicolumn{2}{c}{oven-mitts} & \multicolumn{2}{c}{slice-banana} & \multicolumn{2}{c}{split-cookie} & \multicolumn{2}{c}{torchocolate} & \multicolumn{2}{c}{average}                  \\
         &  mIoU      &  mAcc     &  mIoU      &  mAcc    &  mIoU      &  mAcc    &  mIoU      &  mAcc    &  mIoU      &  mAcc  \\
    \midrule
    SA4D &66.09 &91.30  &77.25  &91.21  &86.40  &99.32  &80.89  &99.47  &81.58  &96.89\\
    SADG &92.18 &98.63  &89.36  &96.58  &86.17  &99.32  &\textbf{87.56}  &\textbf{99.70}  &86.39  &98.45\\
    Our &\textbf{92.89}  &\textbf{98.84}  &\textbf{90.08}  &\textbf{96.92}  &\textbf{93.74}  &\textbf{99.73}  &86.32  &99.68  &\textbf{90.49}  &\textbf{98.80}\\
    \bottomrule
  \end{tabular}
  \label{tab:hypernerf_result}
\end{table}

\begin{table}
 \caption{
 Quantitative comparison of our method’s performance against SA4D \cite{ji2024segment} and SADG \cite{li2024sadg} on the Neu3D \cite{li2022neural} dataset. Our method outperforms these SOTA methods on all scenes (both mIoU and mAcc in \%).
 }
  \centering
  \begin{tabular}{lllllccllllll}
    \toprule
    \multirow{2}{*}{Methods}  & \multicolumn{2}{c}{coffee\_martini} & \multicolumn{2}{c}{cook\_spinach} & \multicolumn{2}{c}{cut\_roasted\_beef} & \multicolumn{2}{c}{flame\_steak} & \multicolumn{2}{c}{sear\_steak} & \multicolumn{2}{c}{average}                   \\
        &   mIoU     &  mAcc     &   mIoU     &  mAcc    &   mIoU      &  mAcc    &   mIoU      &  mAcc    &   mIoU     &  mAcc    &   mIoU     &  mAcc  \\
    \midrule
    SA4D &88.17 &99.28  &88.76  &99.34  &88.98  &99.33  &85.23  &99.16  &89.32  &99.39 &88.09   &99.30\\
    SADG &\textbf{91.47}    &99.50  &91.92  &99.56  &92.88  &99.59  &87.45  &99.32  &89.62  &99.42 &90.67   &99.48\\
    Our &91.28  &\textbf{99.51} &\textbf{94.44} &\textbf{99.70} &\textbf{92.99} &\textbf{99.62} &\textbf{91.38} &\textbf{99.56} &\textbf{93.31} &\textbf{99.63} &\textbf{92.68} &\textbf{99.60}\\
    \bottomrule
  \end{tabular}
  \label{tab:neu3d_result}
\end{table}

\begin{figure}
  \centering
  \includegraphics[width=0.9\textwidth]{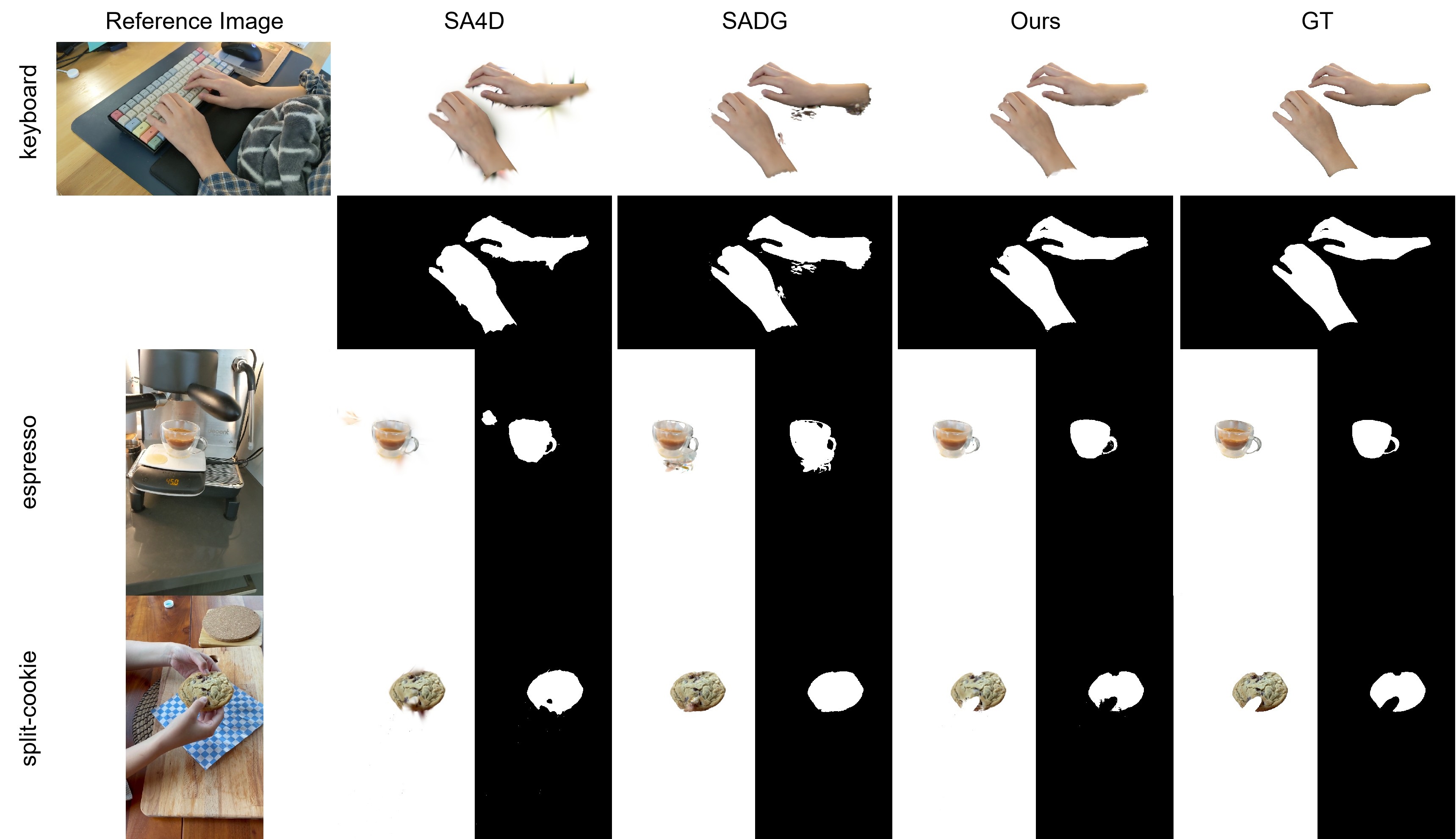}
  \caption{
    Qualitative comparison of our method’s performance against SA4D \cite{ji2024segment} and SADG \cite{li2024sadg} on the HyperNeRF \cite{park2021hypernerf} dataset.
  }
  \label{fig:compare_hypernerf}
\end{figure}

\begin{figure}
  \centering
  \includegraphics[width=0.9\textwidth]{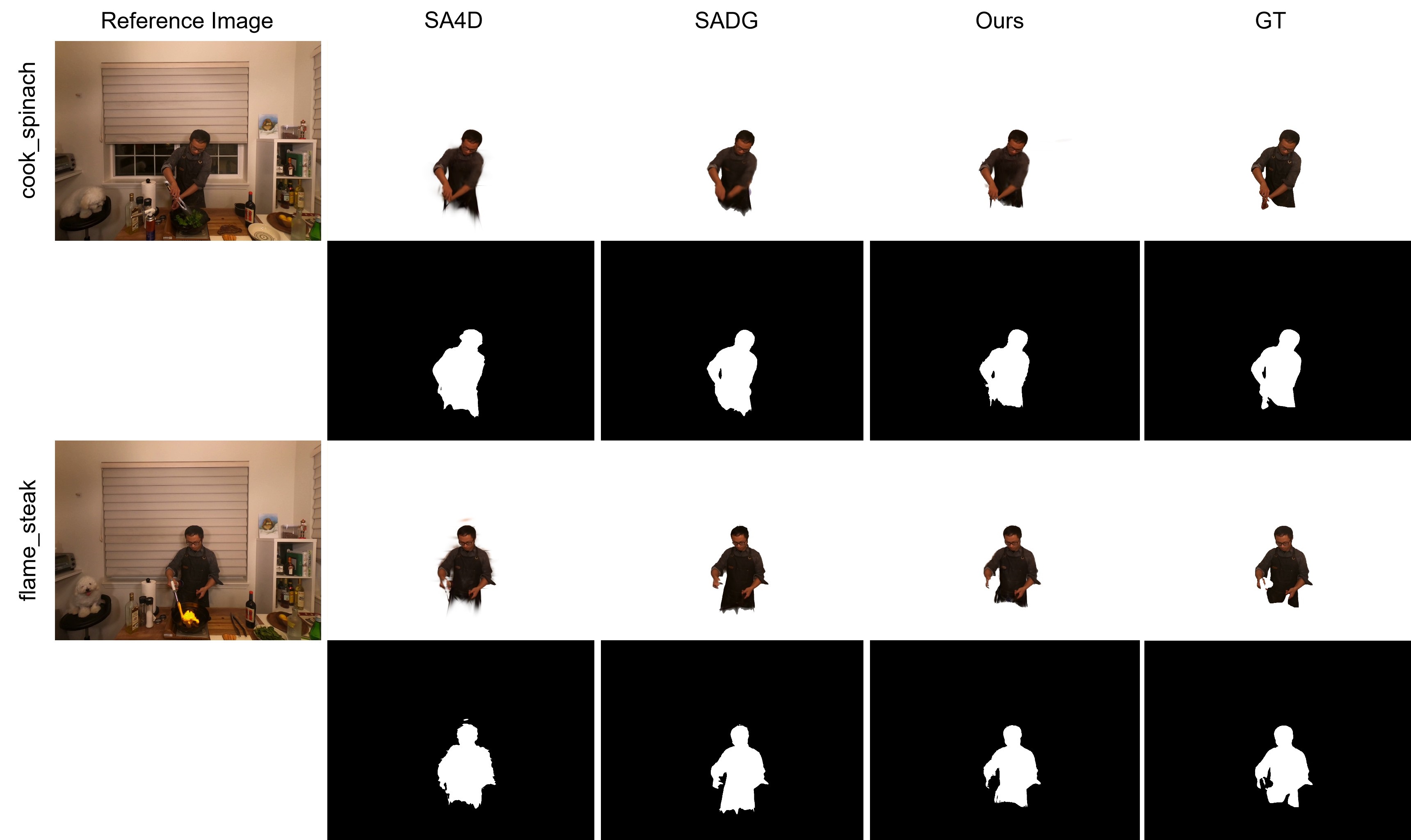}
  \caption{
    Qualitative comparison of our method’s performance against SA4D \cite{ji2024segment} and SADG \cite{li2024sadg} on the Neu3D \cite{li2022neural} dataset.
  }
  \label{fig:compare_neu3d}
\end{figure}


\begin{table}
 \caption{
 Reference time and performance comparison of our method against SA4D \cite{ji2024segment} and SADG \cite{li2024sadg} on HyperNeRF \cite{park2021hypernerf} and Neu3D \cite{li2022neural} dataset. Time denotes the average time (in minutes) required to obtain the target object on each dataset, measured as the total time spent on identity feature learning and target object Gaussians extraction. For our method, we report results with 10 / 5 / 3 / 1 iterations for both IGIT and RCC stages.
 }
  \centering
  \begin{tabular}{ccccccc}
    \toprule
    \multirow{2}{*}{Methods}  &\multicolumn{3}{c}{HyperNeRF} &\multicolumn{3}{c}{Neu3D}\\
    \cmidrule(r){2-4} \cmidrule(r){5-7} 
     &time(min) &mIoU(\%)  &mAcc(\%)  &time(min) &mIoU(\%)  &mAcc(\%)  \\
    \midrule
    SA4D &19.71 &81.58 &96.89 &19.71 &88.09 &99.30\\
    SADG &63.05 &86.39 &98.45 &93.48 &90.67 &99.48\\
    \midrule
    Ours (10 iterations) &0.99 &\textbf{90.53} &\textbf{98.81} &5.97 &\textbf{92.69} &\textbf{99.60}\\
    Ours (5 iterations) &0.51 &90.49 &\textbf{98.81} &3.01 &92.47 &99.59\\
    Ours (3 iterations) &0.30 &89.63 &98.72 &1.75 &91.55 &99.54\\
    Ours (1 iterations) &\textbf{0.11} &81.95 &97.28 &\textbf{0.60} &76.61 &98.14\\
    \bottomrule
  \end{tabular}
  \label{tab:time_compare}
\end{table}

\subsection{Comparison}
\label{sec:Comparison}
Tables \ref{tab:hypernerf_result} and \ref{tab:neu3d_result} report the quantitative segmentation results of our method compared with SA4D~\cite{ji2024segment} and SADG~\cite{li2024sadg} on the HyperNeRF~\cite{park2021hypernerf} and Neu3D~\cite{li2022neural} datasets, following the benchmark annotations provided in SADG~\cite{li2024sadg}. Our method outperforms the other two methods in terms of both mIoU and mAcc.

For the HyperNeRF~\cite{park2021hypernerf} dataset, we visualize the segmentation results of the \textit{keyboard}, \textit{espresso}, and \textit{split-cookie} scenes in Figure~\ref{fig:compare_hypernerf}. Our method achieves notable improvements over the other two approaches. In the \textit{keyboard} and \textit{espresso} scenes, our method produces cleaner object boundaries both visually and in the mask results, whereas SA4D~\cite{ji2024segment} and SADG~\cite{li2024sadg} exhibit some erroneous floating points, and SA4D~\cite{ji2024segment} additionally shows substantial boundary leakage that negatively affects the visual quality. In the \textit{split-cookie} scene, our method outperforms SADG~\cite{li2024sadg} due to the ability to adapt Gaussian instance identities over time, which allows handling parts of the cookie occluded by fingers. However, in the \textit{cut-lemon1} scene, our method does not achieve better quantitative results than SADG~\cite{li2024sadg} due to errors in DEVA\cite{cheng2023tracking} segmentation under this complex scenario.

For the Neu3D~\cite{li2022neural} dataset, we visualize the segmentation results for the \textit{cook\_spinach} and \textit{flame\_steak} scenes in Figure~\ref{fig:compare_neu3d}. Since we only use training views closest to the test set to generate 2D masks for both our method and SA4D~\cite{ji2024segment}, the lack of multi-view supervision leads to some Gaussians that are invisible from the training views being unhandled. Nevertheless, our method still achieves higher quantitative metrics than SADG~\cite{li2024sadg}, largely because our method better suppresses boundary overflow in occluded lower-body regions affected by foreground objects.

Table \ref{tab:time_compare} reports the reference time required by our method, SA4D~\cite{ji2024segment}, and SADG~\cite{li2024sadg} to extract target objects on the HyperNeRF~\cite{park2021hypernerf} and Neu3D~\cite{li2022neural} datasets, measured as the average per-scene total time for instance feature learning and target object segmentation. For our method, we present results for 10, 5, 3, and 1 iterations in both the iterative GIT stage and the Gaussian rendering range control stage. The runtime scales roughly proportionally with the number of iterations. In practice, most scenes converge within 3–5 iterations, and for simpler scenes, fewer iterations can be used to reduce computation time without noticeably affecting segmentation quality.

\begin{figure}[t]
\centering
\begin{minipage}[t]{0.48\linewidth}
\centering
\vspace{0pt} 
\captionof{table}{\textbf{Ablation study}.We evaluate our method under different configurations on \textit{cut-lemon1} and \textit{chickchicken}.} 
\begin{tabular}{lllll}
\toprule
   \multirow{2}{*}{\quad Methods}  &\multicolumn{2}{c}{\textit{cut-lemon1}} &\multicolumn{2}{c}{\textit{chickchicken}} \\
    \cmidrule(r){2-3} \cmidrule(r){4-5} 
    &mIoU  &mAcc   &mIoU  &mAcc \\
\midrule
    (i) w/o TS\&RRC &83.35	&96.35	&91.42	&97.90 \\
    (ii) w/o TS	&84.63	&97.09	&95.21	&98.89 \\
    (iii) w/o RRC	&83.26	&96.40	&91.41	&97.90 \\
\midrule
    (iv) Full ($\tau$=0.3)	&84.57	&97.07	&95.19	&\textbf{98.89} \\
    (v) Full ($\tau$=0.5)	&85.25	&97.20	&\textbf{95.22}	&\textbf{98.89} \\
    (vi) Full ($\tau$=0.7)	&\textbf{86.21}	&\textbf{97.34}	&95.02	&98.85 \\
\bottomrule
\end{tabular}
\label{tab:Ab_result}
\end{minipage}
\hfill
\begin{minipage}[t]{0.48\linewidth}
\centering
\vspace{0pt} 
\includegraphics[width=\linewidth]{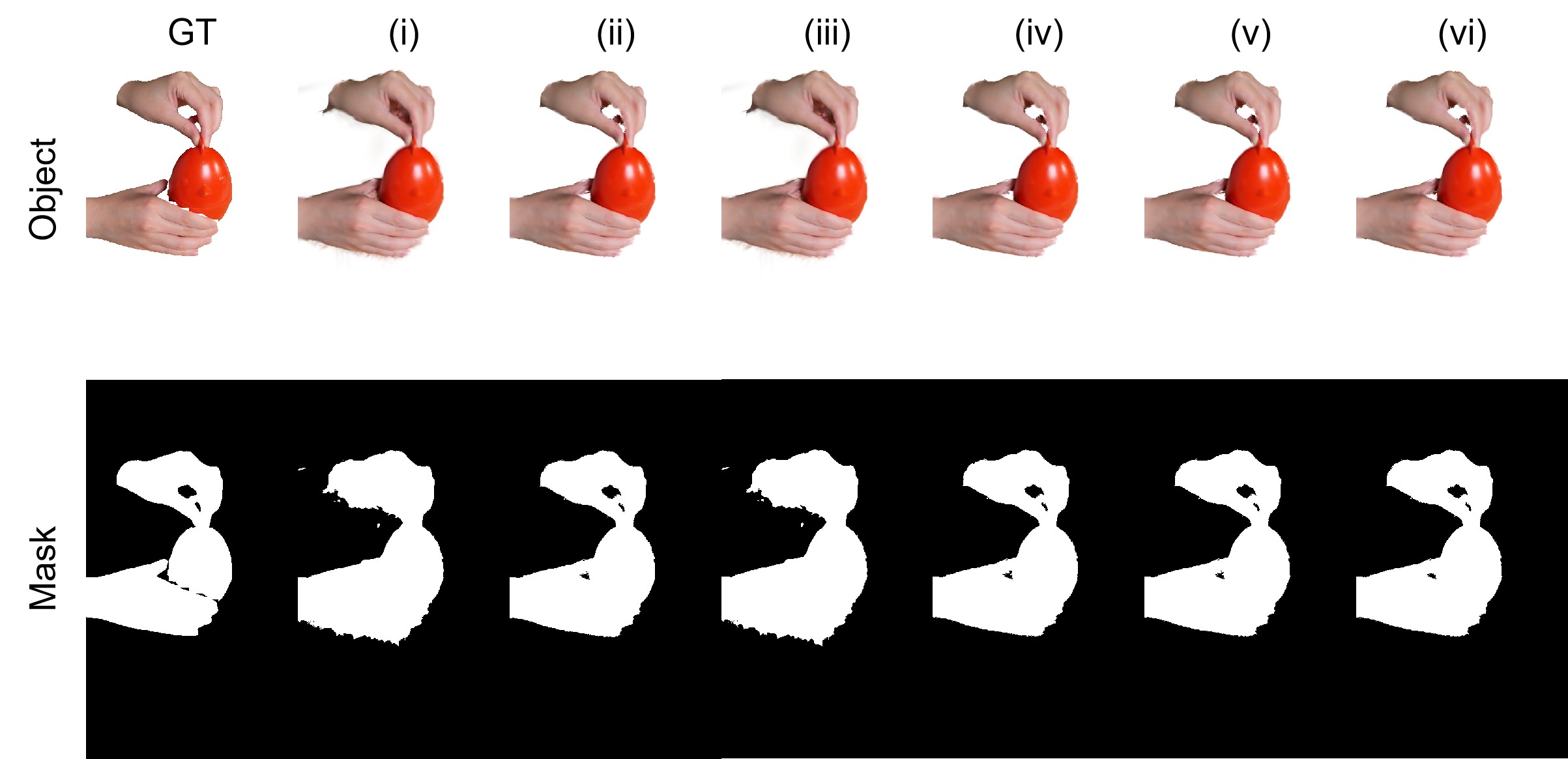}
\captionof{figure}{\textbf{Ablation study}. We present the corresponding qualitative results for each configuration on \textit{chickchicken}.}
\label{fig:Ab_chick}
\end{minipage}
\begin{minipage}[t]{0.96\linewidth}
  \centering
  \vspace{1pt} 
  \includegraphics[width=\textwidth]{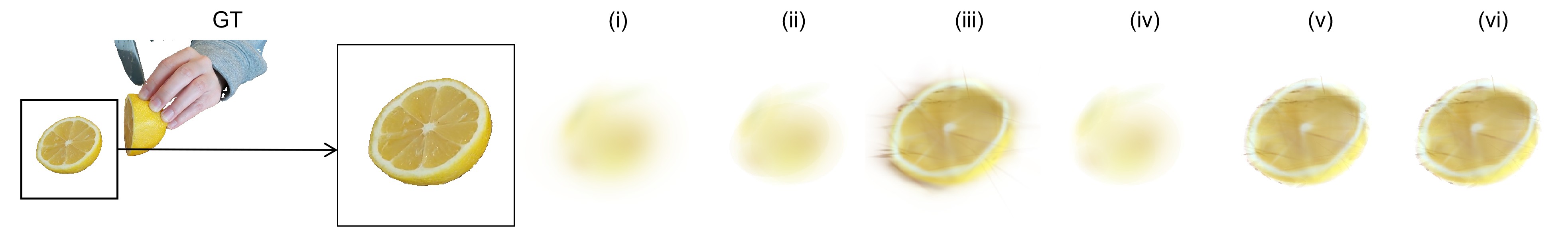}
  \caption{
   \textbf{Ablation study}. We present the corresponding qualitative results for each configuration on \textit{cut-lemon1}.
  }
  \label{fig:Ab_lemon}
\end{minipage}
\end{figure}


\subsection{Ablation Studies}
\label{sec:Ablation}
Table \ref{tab:Ab_result} presents an ablation study analyzing the contributions of different components of our method. We evaluate three variants by disabling the temporal segmentation module (Sec. \ref{sec:TSM}), the Gaussian rendering range control module (Sec. \ref{sec:RRC}), and both modules simultaneously. Experiments are conducted on \textit{cut-lemon1}, where the identity of target-object Gaussians changes significantly, and \textit{chickchicken}, which represents a more stable scene where Gaussian identities remain largely consistent over time.
When the Gaussian rendering range control strategy is removed, both mIoU and mAcc drop substantially, and qualitative results (Figure~\ref{fig:Ab_chick}) on the \textit{chickchicken} scene further show that the rendering range control strategy effectively suppresses Gaussian overflow in boundary regions, indicating that this strategy plays a critical role in improving segmentation accuracy.

The impact of the temporal segmentation strategy on quantitative performance depends on the stability of Gaussian instance identities in the scene. For scenes where Gaussian identities change frequently over time, introducing temporal segmentation with a relatively high merging threshold $\tau$ leads to improved results. In contrast, for scenes with largely stable Gaussian identities, temporal segmentation has only a minor effect on the metrics, and an inappropriately large $\tau$ may leads to overly fine-grained temporal segmentation, resulting in excessively short temporal segments, reducing multi-view constraints within each segment and consequently degrading performance. In terms of qualitative results, without temporal segmentation, the method fails to correctly handle Gaussians that belong to the target objects only during limited temporal intervals due to identity changes, resulting in inferior segmentation in dynamic scenes such as \textit{cut-lemon1} (Figure~\ref{fig:Ab_lemon}).

\subsection{Limitations}
\label{sec:limit}
Though TIBR4D can achieve efficient 4D Gaussian segmentation with high-quality boundaries, several limitations exist and can be explored in the future: 
1) Instance tracing heavily depends on the accuracy of 2D instance masks. Although the proposed two iterative stages substantially reduce spurious floating Gaussians and boundary leakage, severe errors in the 2D segmentation may still propagate into 4D Gaussians and are difficult to eliminate. 
2) Our method focuses on iterative refinement for target object extraction and is therefore not intended for full panoptic segmentation. Although panoptic maps can be obtained either by directly projecting instance tracing probabilities or by extracting and merging individual objects, both strategies are suboptimal in terms of segmentation quality or computational efficiency. 
3) Though our method is convergent, increasing the number of iterations or training views inevitably increases runtime. In practice, for scenes with dense view coverage and limited temporal variation, a downsampling of views usually provides an effective trade-off between efficiency and segmentation accuracy.


\section{Conclusion}
\label{sec:conclu}
We presented an object-level segmentation framework for dynamic Gaussian scenes by lifting video segmentation results into 4D scenes, without requiring additional training. The core of this framework is a two-stage tracing-guided iterative boundary refinement, TIBR4D. 
In the first stage, the input video is first divided into segments according to temporal variations of Gaussian identity features. Then, Iterative Gaussian Instance Tracing (IGIT) is proposed and applied on these temporal segments, rather than globally or locally as existing methods. Thus, TIBR4D is able to extract cleaner target object Gaussian point clouds with fewer erroneous Gaussians, while effectively handling Gaussians whose instance identities vary over time. 
In the second stage, the iterative frame-wise Gaussian Rendering Range Control (RCC) is proposed to avoid Gaussians suffering from boundary overflow or unstable identity assignments within a temporal segment. By restricting each Gaussian to its most reliable contribution region, this strategy produces more accurate object boundaries while avoiding excessive sparsity in the segmented point cloud. 
%
Extensive experiments on the HyperNeRF~\cite{park2021hypernerf} and Neu3D~\cite{li2022neural}  benchmarks demonstrate that our method achieves superior segmentation accuracy compared to existing 4D Gaussian segmentation approaches~\cite{ji2024segment,li2024sadg}, while maintaining competitive computational efficiency. Ablation studies further validate the importance of each component, particularly in challenging scenes with boundary overflow and instance identity changes.

\bibliographystyle{unsrt}  
\bibliography{references}

\end{document}